\documentclass{article}

\usepackage{arxiv,times}

\usepackage{url,hyperref,lineno,microtype,subcaption}
\usepackage{amsmath}
\usepackage{graphicx}

\graphicspath{ {.} }

\iclrfinalcopy

\title{Holdout-Based Fidelity and Privacy Assessment of Mixed-Type Synthetic Data}

\author{Michael Platzer \\
  MOSTLY AI \\
  Vienna, Austria \\
  \texttt{michael.platzer@mostly.ai} \\
   \And
  Thomas Reutterer \\
  Vienna University of Economics and Business \\
  Vienna, Austria \\
  \texttt{thomas.reutterer@wu.ac.at} \\
}

\begin{document}
\maketitle

\begin{abstract}
AI-based data synthesis has seen rapid progress over the last several years, and is increasingly recognized for its promise to enable privacy-respecting high-fidelity data sharing. However, adequately evaluating the quality of generated synthetic datasets is still an open challenge. We introduce and demonstrate a holdout-based empirical assessment framework for quantifying the fidelity as well as the privacy risk of synthetic data solutions for mixed-type tabular data. Measuring fidelity is based on statistical distances of lower-dimensional marginal distributions, which provide a model-free and easy-to-communicate empirical metric for the representativeness of a synthetic dataset. Privacy risk is assessed by calculating the individual-level distances to closest record with respect to the training data. By showing that the synthetic samples are just as close to the training as to the holdout data, we yield strong evidence that the synthesizer indeed learned to generalize patterns and is independent of individual training records. We demonstrate the presented framework for seven distinct synthetic data solutions across four mixed-type datasets and compare these to more traditional statistical disclosure techniques. The results highlight the need to systematically assess the fidelity just as well as the privacy of these emerging class of synthetic data generators.
\end{abstract}

\section{Introduction}

Self-supervised generative AI has made significant progress over the past years, with algorithms capable of creating “shockingly” realistic synthetic data across a range of domains. Demonstrations like are particularly impressive within domains of unstructured data, like images \citep{karras2017progressive} and text \citep{brown2020language}. These samples demonstrate that it is becoming increasingly difficult for us humans, as well as for machines, to discriminate actual from machine-generated fake data. While less prominent, similar progress is made within structured data domains, such as synthesizing medical health records \citep{choi2017generating, krauland2020development, goncalves2020generation}, census data \citep{freiman2017data}, human genoms \citep{yelmen2021creating}, website traffic \citep{lin2020using} or financial transactions \citep{assefa2020generating}. These advances are particularly remarkable considering that they do not build upon our own human understanding of the world, but “merely” require a flexible, scalable self-supervised learning algorithm that teaches itself to create novel records based on a sufficient amount of training data.

Given this new capability to generate arbitrary amounts of new data, many applications arise and provide rich opportunities. These range from automated content creation \citep{shu2020fact}, test data generation \citep{popic2019data}, world simulations for accelerated learning \citep{ha2018world}, to general-purpose privacy-safe data sharing \citep{Surendra2017ARO, howe2017synthetic, bellovin2019privacy, hittmeir2019utility, li2019evaluating}.

We focus on the data sharing use cases, where data owners seek to provide highly accurate, yet truly anonymous statistical representations of datasets. AI-based approaches for generating synthetic data provide a promising novel tool box for data stewards in the field of statistical disclosure control (SDC) \citep{drechsler2011synthetic}, but just as more traditional methodologies also share the fundamental need to balance data utility against disclosure risk. One can maximize utility by releasing the full original dataset, but would thereby expose the privacy of all contained data subjects. On the other hand, one can easily minimize the risk by releasing no data at all, which naturally yields zero utility. It is this privacy-utility trade-off that we seek to quantify for mixed-type synthetic data by introducing the empirical assessment framework proposed in this paper. After briefly discussing the background we present the building blocks of the proposed framework in section \ref{sec:methods}. This will then allow us to compare the performance of generative models from the rapidly growing field of synthetic data approaches against each other, as well as against alternative SDC techniques in section \ref{sec:results}.

\section{Related Work}\label{sec:related}

The field of generative AI gained strong momentum ever since the introduction of Generative Adversarial Networks \citep{goodfellow2014generative} and its application to image synthesis. This seminal and widely cited paper assessed synthetic data quality by fitting Gaussian Parzen windows to the generated samples in order to estimate the log-likelihood of holdout samples. At that time the authors already called out for further research to assess synthetic data, as they highlighted the limitations of Parzen window estimates for higher dimensional domains, which are then also further confirmed by \cite{theis2015note}. In addition to quantitative assessments, nearly all of the research advances for image synthesis also present non-cherry picked synthetic samples as an indicator for quality (see e.g., \citealt{karras2017progressive, liu2016coupled, radford2015unsupervised}). While these allow to visually judge plausibility of the generated data, they do not allow to capture a generator’s ability to faithfully represent the full variety and richness of a dataset, i.e. its dataset-level statistics. On the contrary, by overly focusing on “realistic” sample records in the assessment, one will potentially favor generators that bias towards conservative, safe-bet samples, at the cost of diversity and representativeness.\footnote{Temperature-based sampling \citep{ackley1985learning}, top-k sampling \citep{fan2018hierarchical}, and nucleus sampling \citep{holtzman2019curious} are all techniques to make such trade-offs explicitly, and are commonly applied for synthetic text generation.}

For structured data a popular and intuitive approach is to visually compare histograms and correlation plots (see e.g., \citealt{lu2019empirical, howe2017synthetic}). While this does allow to capture representativeness, it typically is being applied to only a small subset of statistics, and misses out on systematically quantifying any discrepancies thereof.

A popular assessment technique common to both structured as well as unstructured domains is to train a supervised machine learning task on the generated synthetic data and see how its predictive accuracy fairs against the same model being trained on real data \citep{jordon2018measuring, xu2019modeling}. By validating against an actual holdout dataset, that is not used for the data synthesis itself, one gets an indication for the information loss for a specific relationship within the data incurred due to the synthesis. If the chosen predictive task is difficult enough and a capable downstream machine learning model is used, this can indeed yield a strong measure. However, results will depend on both of these assumptions, and will vary even for the same dataset from predicted target to predicted target, as it tests only for a singular relationship within the high dimensional data distribution. And more importantly, the measure again does not allow statements regarding the overall statistical representativeness. Any accidentally introduced bias, any artefacts, any misrepresentations within the generated data might remain unnoticed. Yet, all of these are of particular importance when a data owner seeks to disseminate granular-level information with highest possible accuracy, without needing to restrict or even to know the downstream application.

No accuracy assessment of a synthetic data solution can be complete, if it does not include some measurement of its ability to produce truly novel samples, rather than merely memorizing and recreating actual data. Closely related, users of synthetic data solutions seek to establish the privacy of a generated dataset. I.e. whether the synthetic dataset is considered to be anonymous, non-personal data in a legal sense. With data protection regulations varying from country to country, and industry to industry, any ultimate assessment requires legal expertise and can only be done with respect to a given regulation. However, there are a growing number of technical definitions and assessments of privacy being introduced, that serve practitioners well to make the legal case. Two commonly used concepts within the context of synthetic data are empirical attribute disclosure assessments \citep{hittmeir2020baseline, taub2018differential}, and Differential Privacy \citep{dwork2006calibrating}. Both of these have proven to be useful in establishing trust in the safety of synthetic data, yet come with their own challenges in practice. While the former requires computationally intensive, case-specific repeated synthetization re-runs that can become infeasible to perform on a continuous base, the latter requires the inspection of the algorithms as well as their actual implementations for these to be validated. We seek to contribute to the growing list of privacy concepts by proposing an easy-to-compute holdout-based empirical assessment, that does not rely on knowledge of the underlying synthetization process.

\section{Framework}\label{sec:methods}

\subsection{Fidelity}

\begin{figure}[ht]
\begin{center}
\includegraphics[width=13.5cm]{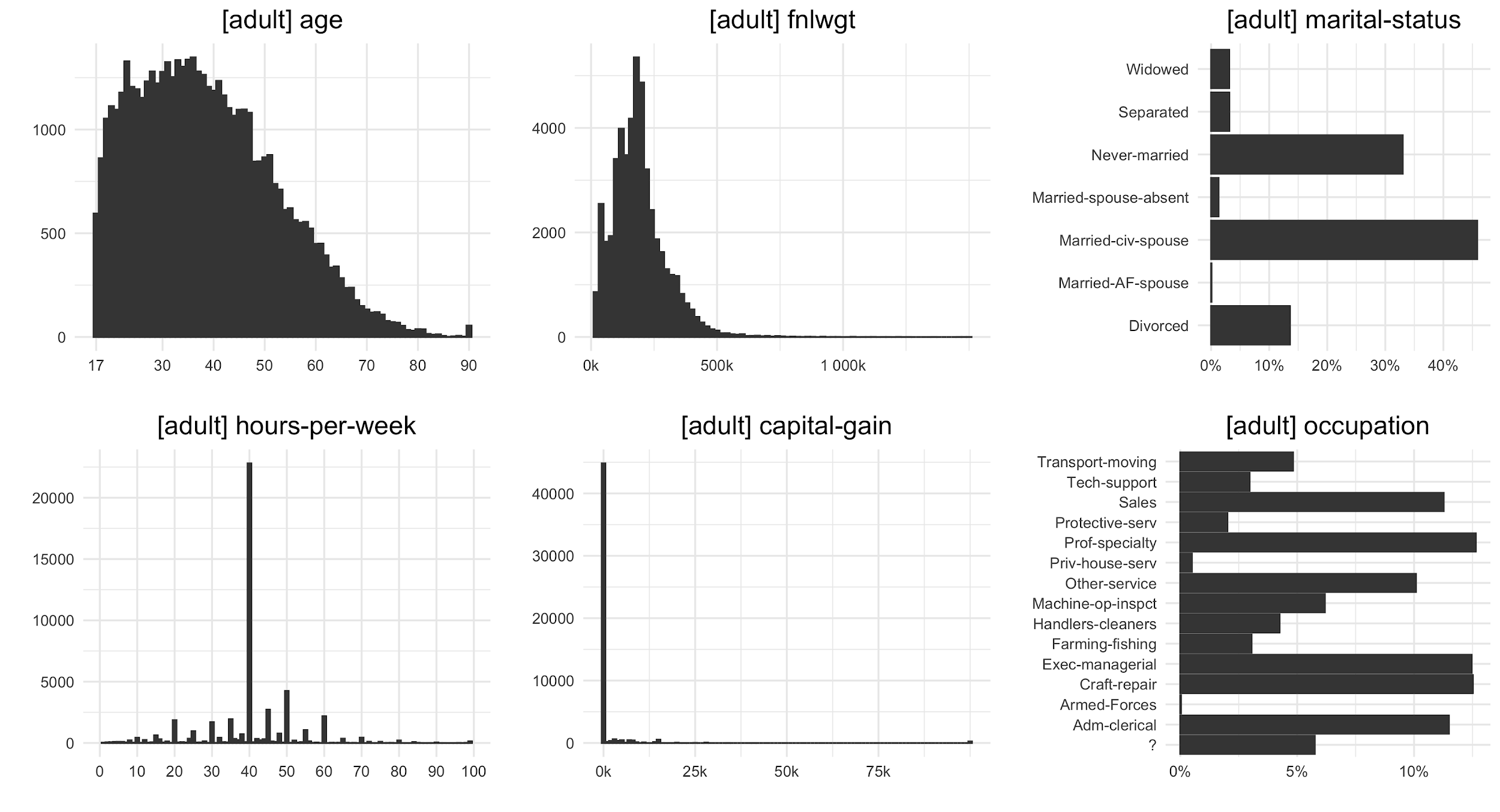}
\end{center}
\caption{Selected univariate marginal distributions for dataset `adult`}\label{fig:adult-univariate}
\end{figure}

We start out by motivating our introduced fidelity measure by visualizing selected distributions and cross-tabulations for the `adult` dataset, which we will use later in our empirical demonstration study. Figure \ref{fig:adult-univariate} exhibits the distribution of four selected numeric attributes and shows the wide variety of shapes that can occur in real-world datasets. For example, the numeric attribute `age` ranges from 17 and 90, with a small cluster of subjects that are exactly 90 years old, while hardly any subject is between 85 and 89 years old. The Numeric attribute `fnlwgt` spans a much wider range, with nearly all observed values being unique within the dataset. Thus these values need to be binned to visualize the shape of the variable's distribution. Attribute `hours-per-week` is characterized by specific outstanding integer values, while `capital-gain` is dominated by zeros with only a few exceptions that themselves can range up to 100’000. We would want to see synthesizers faithfully retain any of these different types and shapes of univariate distributional patterns and thus need a fidelity measure that can capture any discrepancies thereof.

\begin{figure}[ht]
\begin{center}
\includegraphics[width=13.5cm]{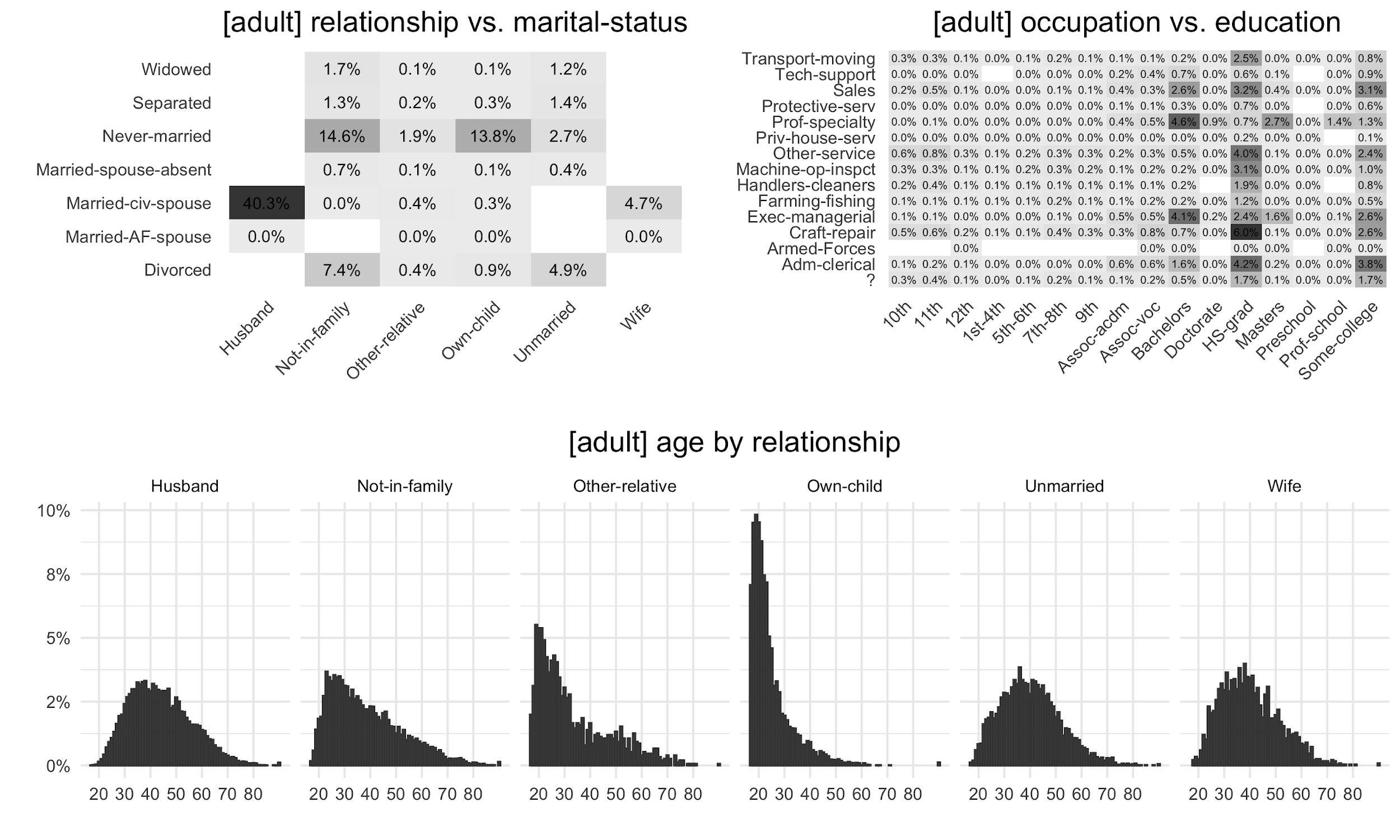}
\end{center}
\caption{Selected bivariate marginal distributions for dataset `adult`}\label{fig:adult-bivariate}
\end{figure}

However, synthetic data shall not only be representative for the distribution of individual attributes, but for all multivariate combinations and relationships among the set of attributes. For example, figure \ref{fig:adult-bivariate} displays three selected bivariate distributions for the dataset `adult`, each with distinct patterns and insights that are to be retained and assessed. The challenge for deriving a metric that accommodates these interdependencies in an adequate way is that the number of relationships to investigate grows quickly with the number of attributes. More specifically, a dataset with $m$ attributes results in $\binom{m}{k}$ combinations of $k$-way interactions. For example, for 50 attributes this yields 1’225 two-way, and 19’600 three-way interactions. Ideally, we would want to compare the full joint empirical distributions ($m=k$) between the actual and synthetic data, but that is, except for the most trivial cases, infeasible in practice. The curse of dimensionality \citep{bellman1966dynamic} strikes here again; i.e., the number of cross-combinations of attribute values grows exponentially as more attributes are considered, resulting in the available data becoming too sparse in a high-dimensional data space. While binning and grouping of attribute values mitigates the issue for the lower-level interactions, this fundamental principle cannot be defeated for deeper levels. Thus we propose as a practical, model- and assumption-free approach to empirically measure the fidelity of a synthetic dataset with respect to a target dataset by averaging across the total variation distances\footnote{The total variation distance equals half the $L_1$-distance for the discretized distributions at hand. We explored other distance measures for empirical distributions, like the maximum absolute error, euclidean distances, the Jensen-Shannon distance or the Hellinger distance, but they yielded practically identical rankings for the empirical benchmarks. However, the TVD is easy to communicate, easy to reason about and has exhibited in our experiments a low sensitivity with respect to sampling noise.} (TVD) of the corresponding discretized, lower-level empirical marginal distributions.

The construction of our proposed fidelity metric proceeds as follows: Let’s consider a random split of the available records into a training dataset $T$ and a holdout dataset $H$. We only expose the training data $T$ to a synthesizer which yields a synthetic dataset $S$ of arbitrary size $n_S$. Further, let’s transform each of the attributes of these datasets into categorical variables, that have a fixed upper bound $c$ for their cardinality. For those categorical variables that have cardinality $c_j > c$, we lump together the $(c_j - c + 1)$ least frequent values into a single group. For numeric variables we apply quantile binning, i.e., cut the range of values into a maximum of $c$ ranges, based on their $c$ quantiles. Any date or datetime variable is to be converted first into a numeric representation before applying the same transformation as suggested above. Any missing values are treated as yet another categorical value, thus can increase cardinality to $c+1$ for those variables that contain missing values. Note, that the required statistics for the discretization, i.e., the list of least frequent values as well as the quantiles, are to be determined based on the training dataset $T$ alone, and then reused for the discretization of the other datasets.

We then proceed in calculating relative frequencies for all $k$-way interactions for the discretized $m$ attributes and do so for both the training dataset $T$ and the synthetic dataset $S$. For each $k$-way interaction we calculate the TVD between the two corresponding empirical marginal distributions and then average across all \(\binom{m}{k}\) combinations. This yields a measure $F^k(T, S)$, which quantifies the fidelity of dataset $S$ with respect to dataset $T$. In order to help gauge how much information is lost due to the synthetization and how much discrepancies is expected due to sampling noise we shall compare $F^k(T,S)$ with $F^k(T,H)$. The fidelity measure of the holdout with respect to the training data thus serves us as a reference for what we aim for when retaining statistics, that generalize beyond the individuals.

\subsection{Privacy}

While fidelity is assessed at the dataset-level, we need to look at individual-level distances for making the case that none of the training subjects is exposed by any of the generated synthetic records.

A simplistic approach is to check for identical matches, i.e., records from the training set that are also contained in the synthetic set. However, the occurrence of identical matches is neither a required nor a sufficient condition for detecting a leakage of privacy. Just as any dataset can contain duplicate records, we shall expect a similar relative occurrence within a representative synthetic dataset. Further, and analogous to that metaphorical monkey typing the complete works of William Shakespeare by hitting random keys on a typewriter for an infinite time (also known as the `infinite monkey theorem`), even an uninformed random data generator will eventually end up generating any data record. More importantly, these identical matches must not be removed from the synthetic output, as such a rejection filter actually leaks privacy, since it would reveal the presence of a specific record in the training data by it being absent from a sufficiently large generated synthetic dataset.

The concept of identical matches is commonly generalized towards measuring the distance to closest records (DCR) \citep{park2018data, lu2019empirical}. These are the individual-level distances of synthetic records with respect to their corresponding nearest neighboring records from the training dataset. The distance measure itself is interchangeable, whereas we opt for the Hamming distance applied to the discretized dataset as an easy-to-compute, easy-to-reason distance metric. However, the very same framework can be just as well applied on top of alternative distance metrics, including ones based on more meaningful learned representations of domain-specific embedding spaces. A DCR of 0 corresponds to an identical match. But as argued above, also that metric in itself does not reveal anything regarding the leakage of individual-level information, but is rather a statistic of the data distribution we seek to retain. Therefore, to provide meaning and to facilitate interpretation the measured DCRs need to be put into the context of their expected value, which can be estimated based on an actual holdout dataset.

\begin{figure}[ht]
\begin{center}
\includegraphics[width=13.5cm]{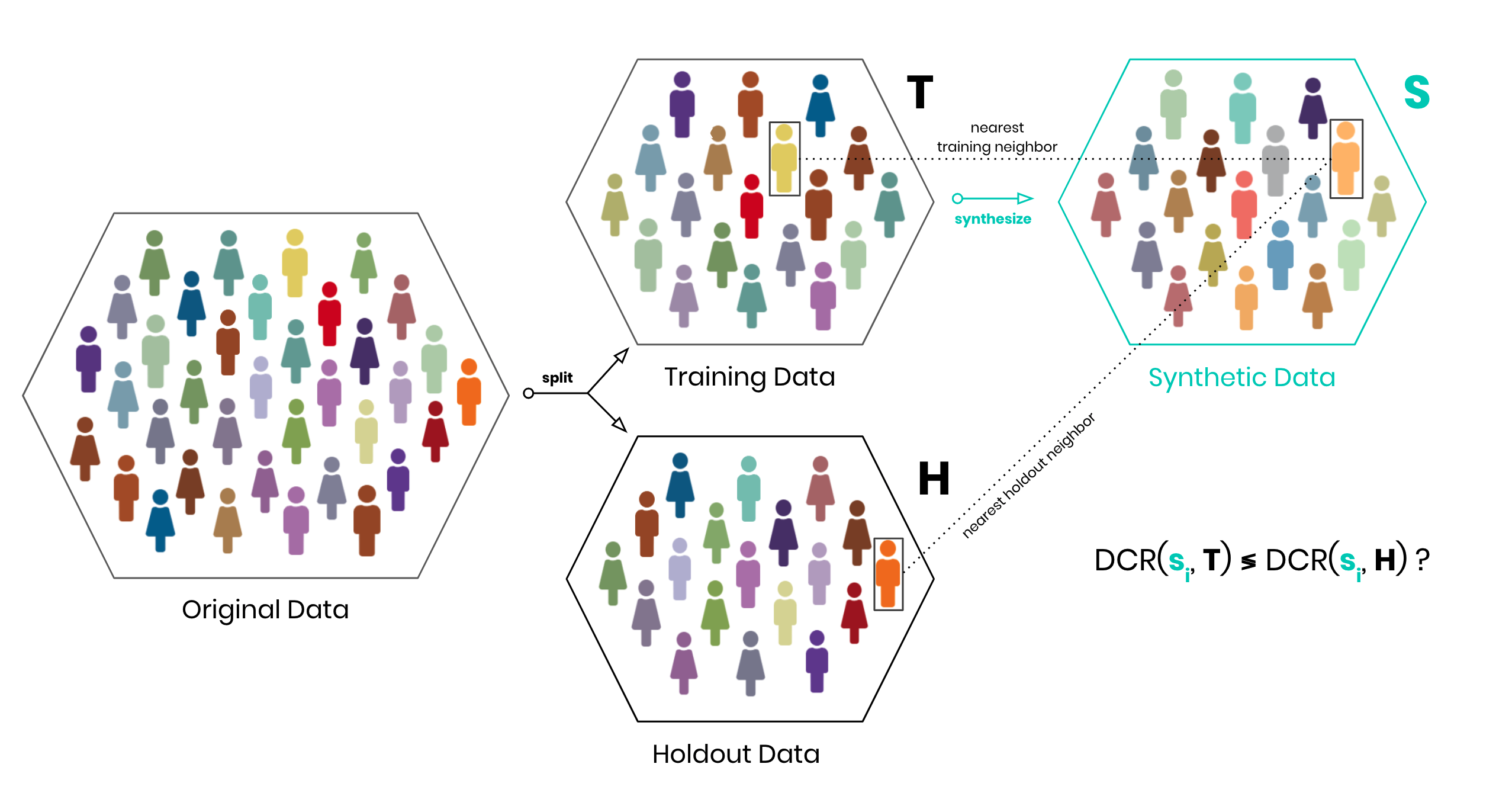}
\end{center}
\caption{Construction of the holdout-based privacy risk measure}\label{fig:framework}
\end{figure}

As illustrated in figure \ref{fig:framework}, we therefore propose to calculate for each synthetic record its DCR with respect to the training data as well as with respect to an equally sized holdout dataset. The \textit{share of records} that are then closer to a training than to a holdout record serves us as our proposed privacy risk measure. Any ties are to be distributed equally between these two datasets. If that resulting share is then close to 50\%, we gain empirical evidence of the training and holdout data being interchangeable with respect to the synthetic data\footnote{Note, that as the holdout records are randomly sampled and never exposed to the synthesizer, the synthesizer can not systematically generate subjects that are closer to these than to the training records. I.e., the presented privacy metric can not be undermined by mixing "too close" with "too far away" records in an attempt to achieve a balanced share.}. This in turn allows to make a strong case for plausible deniability for any individual, as the synthetic data records do not allow to conjecture whether an individual was or was not contained in the training dataset. Even for cases of a strong resemblance of a particular record with a real-world subject, it can be argued that such a resemblance can occur for unseen subjects just as well. Thus, one can deduce that “any resemblance to persons living or dead is purely coincidental”.

\section{Results}\label{sec:results}

To demonstrate the usefulness of the presented framework for assessing fidelity and privacy of synthetic data solutions, we apply it to four publicly available, mixed-type tabular datasets from the UCI Machine Learning repository \citep{Dua2019} and synthesize them using seven publicly available data synthesizers.

The four datasets include:

\begin{itemize}
\item adult: 48’842 records with 15 attributes (6 numerical, 9 categorical)
\item bank-marketing: 45’211 records with 17 attributes (7 numerical, 10 categorical)
\item credit-default: 30’000 records with 24 attributes (20 numerical, 4 categorical)
\item online-shoppers: 12’330 records with 18 attributes (4 numerical, 14 categorical)
\end{itemize}

The seven tested generative models comprise four generators contained as part of MIT’s Synthetic Data Vault (SDV) library \citep{montanez2018sdv}, the synthpop R package \citep{nowok2016synthpop}, an open-sourced generator by Gretel (\href{https://gretel.ai/}{gretel.ai}), plus one closed-source solution by MOSTLY AI (\href{https://mostly.ai/}{mostly.ai}), who provide a freely available community edition online.

Each of the four datasets is randomly split into an equally sized training and holdout dataset. The seven generative models are fitted to the training data to then generate 50'000 synthetic records for each dataset. All synthesizers are run with their default settings unchanged, i.e., no parameter tuning is being performed.

To provide further context to the results, we also generate additional datasets by perturbating the training data with a varying degree of noise. In doing so, we draw 50'000 records with replacement from the dataset and then decide for each value of each record with a given probability (ranging from 10\% up to 90\%) whether to keep it or to replace it with a value from a different record. This approach adds noise to existing records, yet retains the univariate marginal distributions. With more noise being added, one expects privacy to be increasingly protected, while also more statistical relations to be distorted.

\begin{figure}[ht]
\begin{center}
\includegraphics[width=13.5cm]{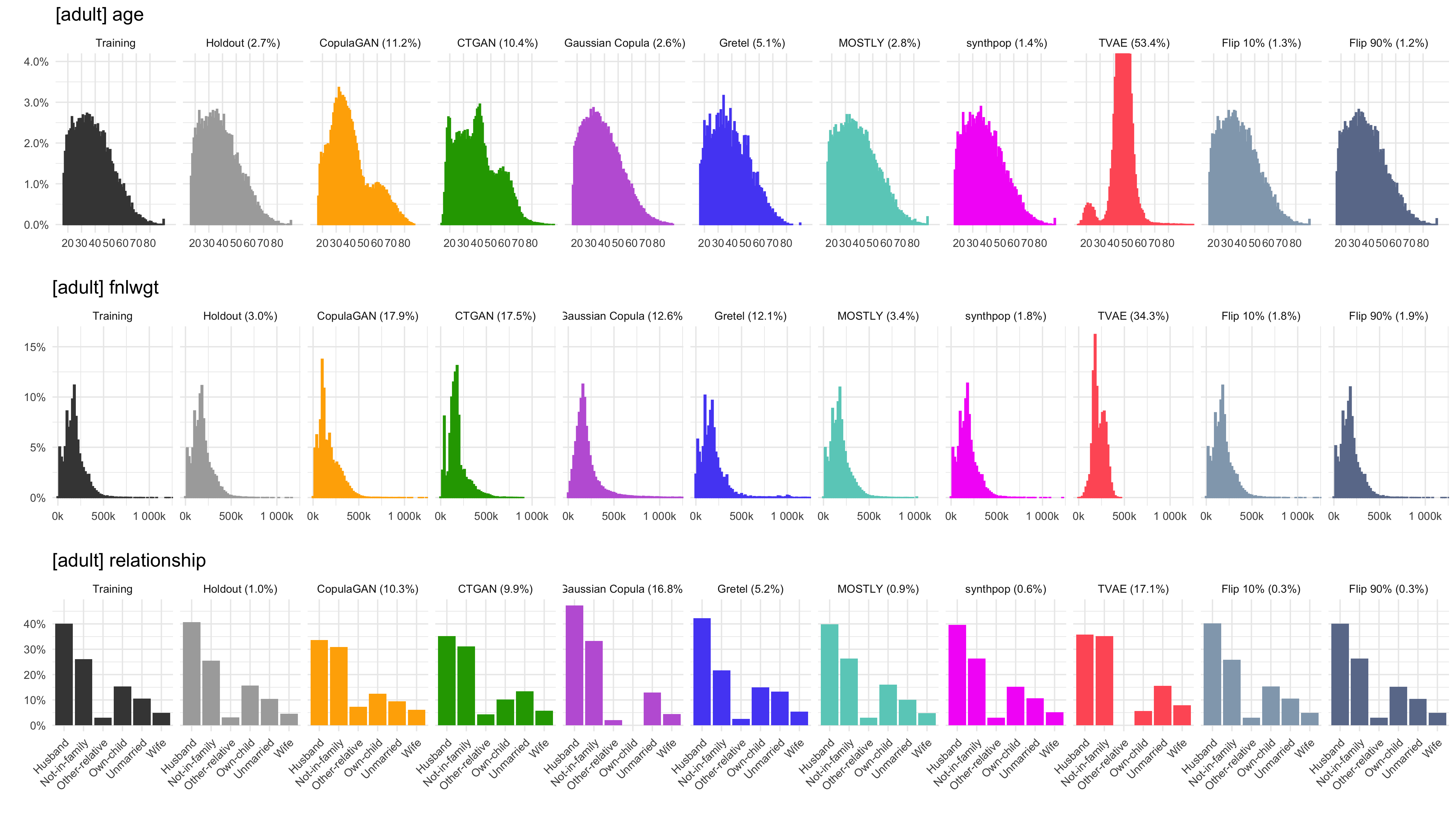}
\end{center}
\caption{Selected univariate marginal distributions for dataset `adult` across generators}\label{fig:adult-univariate-bench}
\end{figure}

\begin{figure}[ht]
\begin{center}
\includegraphics[width=13.5cm]{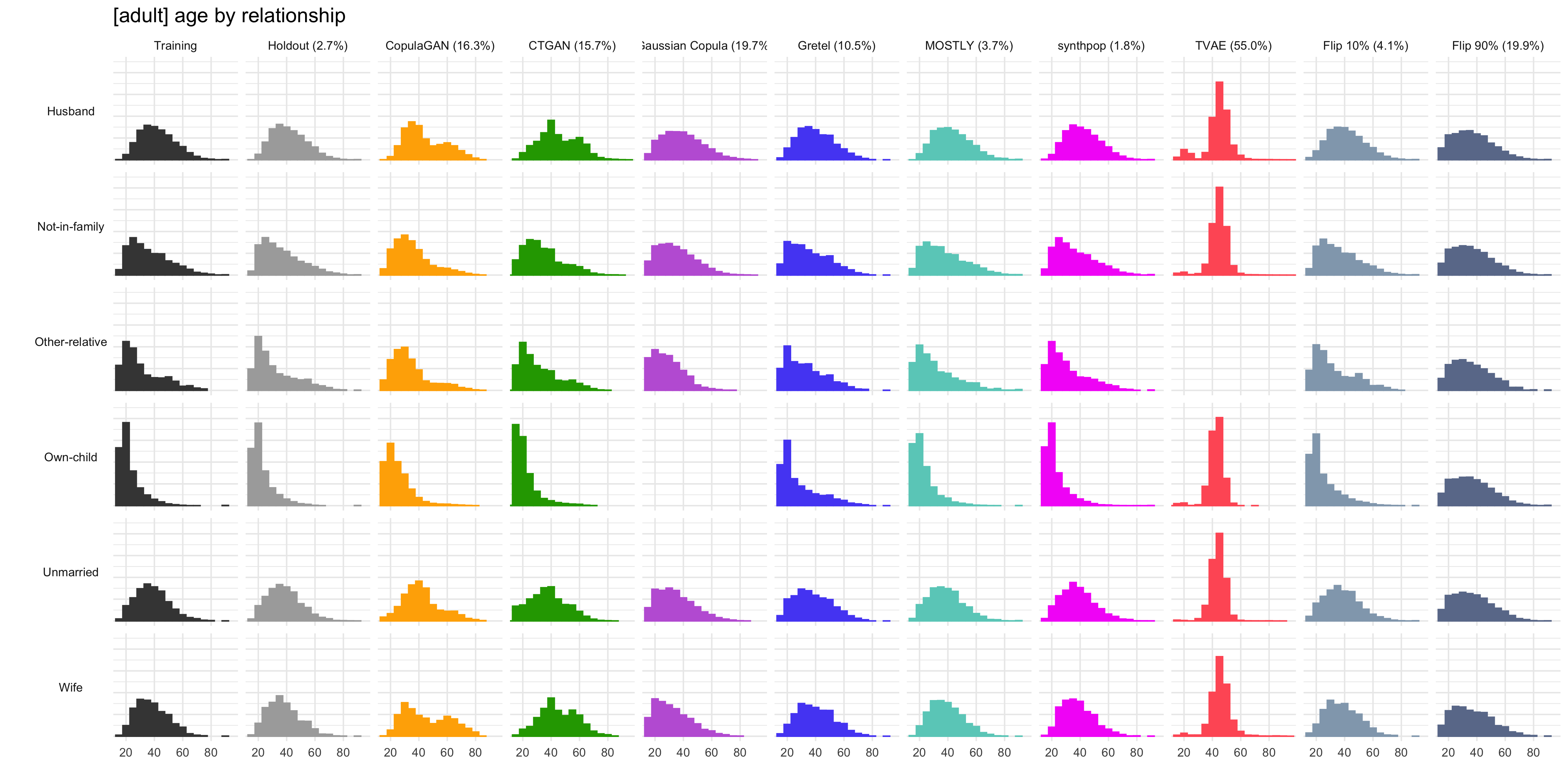}
\end{center}
\caption{ Selected bivariate marginal distribution for dataset `adult` across generators}\label{fig:adult-bivariate-bench}
\end{figure}

\begin{figure}[ht]
\begin{center}
\includegraphics[width=13.5cm]{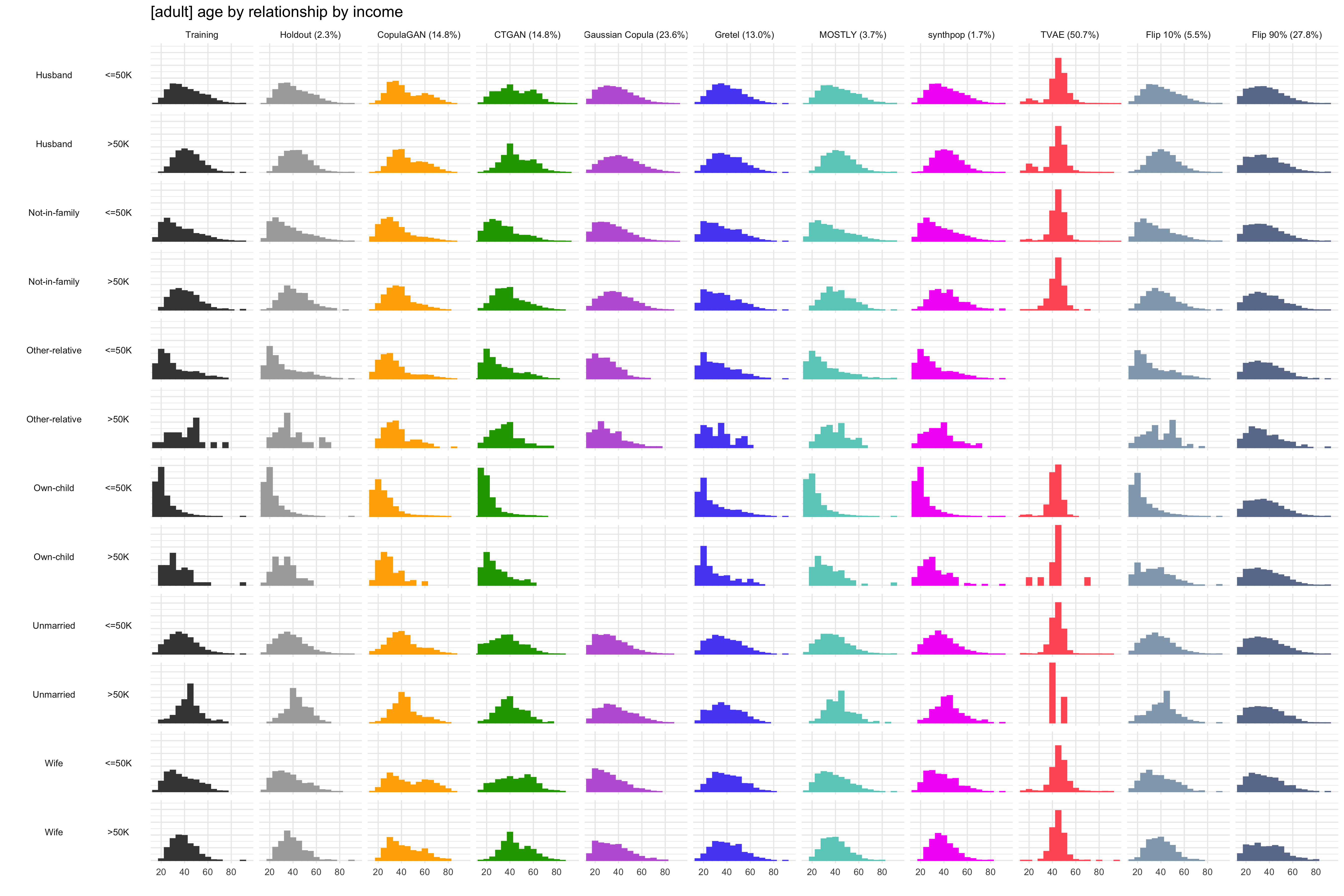}
\end{center}
\caption{Selected three-way marginal distribution for dataset `adult` across generators}\label{fig:adult-threeway-bench}
\end{figure}

\begin{figure}[ht]
\begin{center}
\includegraphics[width=13.5cm]{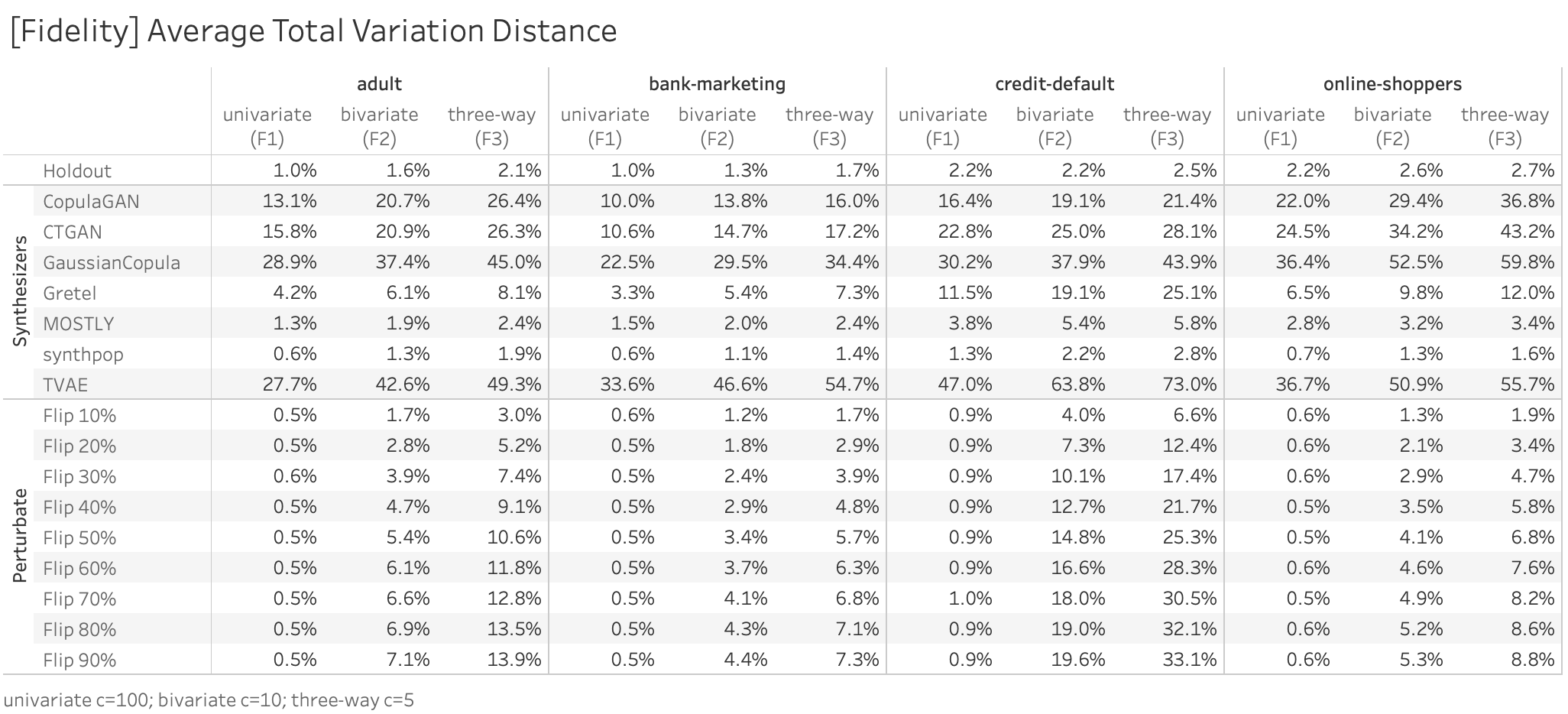}
\end{center}
\caption{Fidelity measures of the presented empirical study}\label{fig:bench-fideltiy}
\end{figure}

Figures \ref{fig:adult-univariate-bench}, \ref{fig:adult-bivariate-bench} and \ref{fig:adult-threeway-bench} visualize the resulting distributions of selected univariate, bivariate and three-way attribute interactions for the `adult` dataset across the various generated datasets. While the visual inspection already allows to spot some qualitative differences with respect to representativeness of the training data, it is the corresponding TVD metric that provides us with a quantitative summary statistic. The reported TVD for the holdout data then serves as a reference, as the derived distributions shall not be systematically closer to the training than what is expected from the holdout. And with 15 univariate, 105 bivariate and 455 three-way interactions for dataset `adult` to be investigated, the proposed summary statistics of averaging across these yield a condensed but informative fidelity assessment of synthetic data. Figure \ref{fig:bench-fideltiy} reports the proposed fidelity measures across all four datasets and generative methods.\footnote{For the fidelity assessment we chose for $c=100$ for discretizing the univariate distributions, $c=10$ for the bivariate combinations, and $c=5$ for the 3-way interactions. Experiments have shown that rankings among synthesizers remain relatively robust with respect to the cardinality of the categorical variable $c$.}

It is interesting to note that the rankings with respect to fidelity among synthesizers are relatively consistent across all datasets, showing that these metrics indeed serve as general-purpose measures for the quality of a synthesizer. The reported numbers for the perturbated datasets exhibit the expected relationship between noise level and fidelity. Among the benchmarked synthesizers `synthpop` and `MOSTLY` exhibit the highest fidelity score with the caveat that the former is systematically too close to the training data compared to what is expected based on the holdout.

\begin{figure}[ht]
\begin{center}
\includegraphics[width=13.5cm]{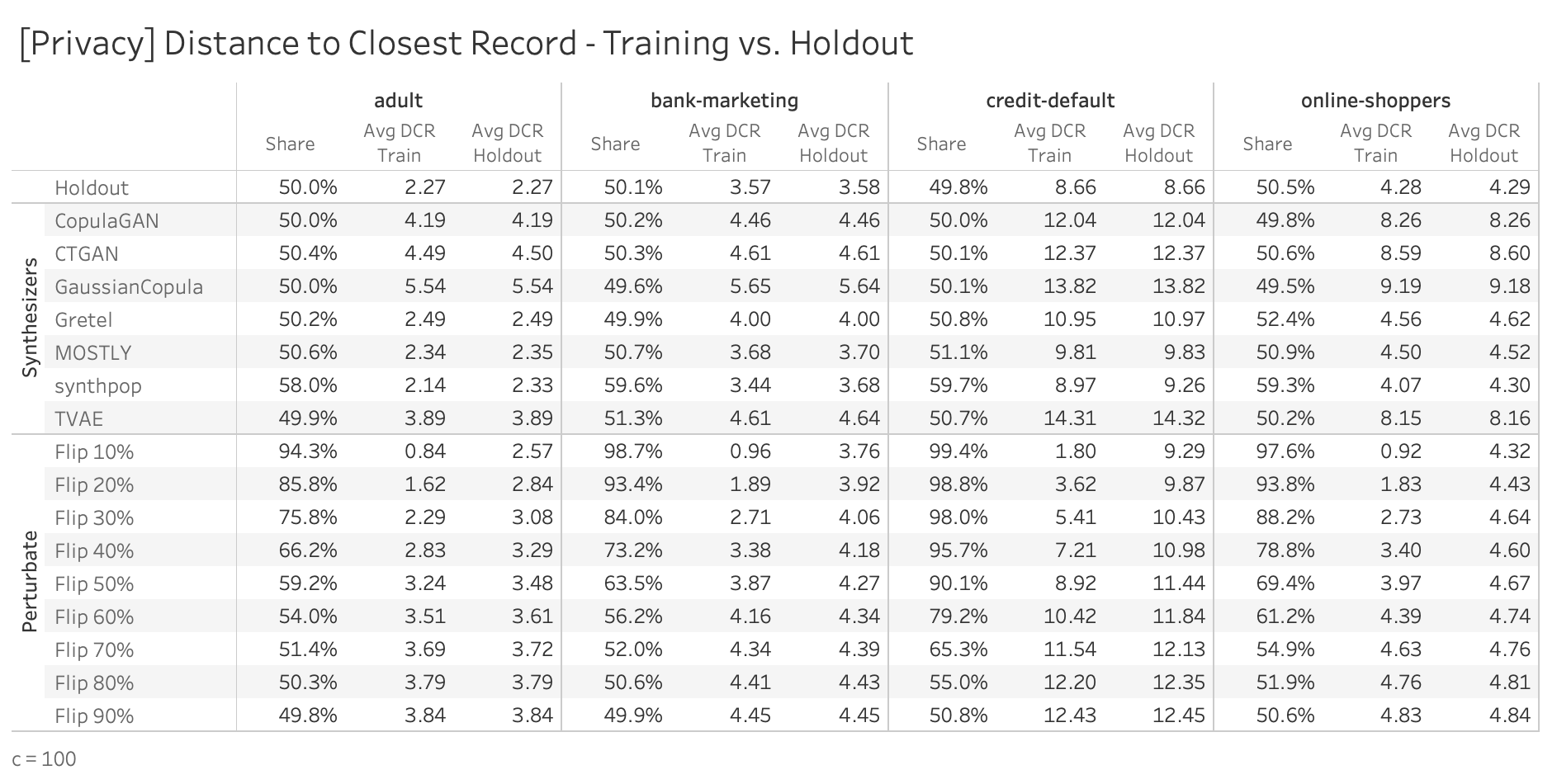}
\end{center}
\caption{Privacy measures of the presented empirical study}\label{fig:bench-privacy}
\end{figure}

Figure \ref{fig:bench-privacy} on the other hand contains the results for the proposed privacy risk measures. For each dataset and synthesizer the share of synthetic records that is closer to a training record than to a holdout record is being reported. In addition, the average DCRs are displayed, once with respect to the training and once with respect to the holdout. With the notable exception of 'synthpop' all of the presented synthesizers exhibit near identical DCR distributions for training as well as for holdout records. This indicates that no individual-level information of the training subjects has been exposed beyond what is attainable from the underlying distribution and thus makes a strong case for the generated data preserving the privacy of the training subjects. In contrast, the reported numbers for the perturbated datasets reveal a severe exposure of the training subjects, even as a high amount of noise is being added. Only at a level where most of the utility of these datasets is being destroyed, the privacy measures start to align with the holdout dataset.

\begin{figure}[ht]
\begin{center}
\includegraphics[width=13.5cm]{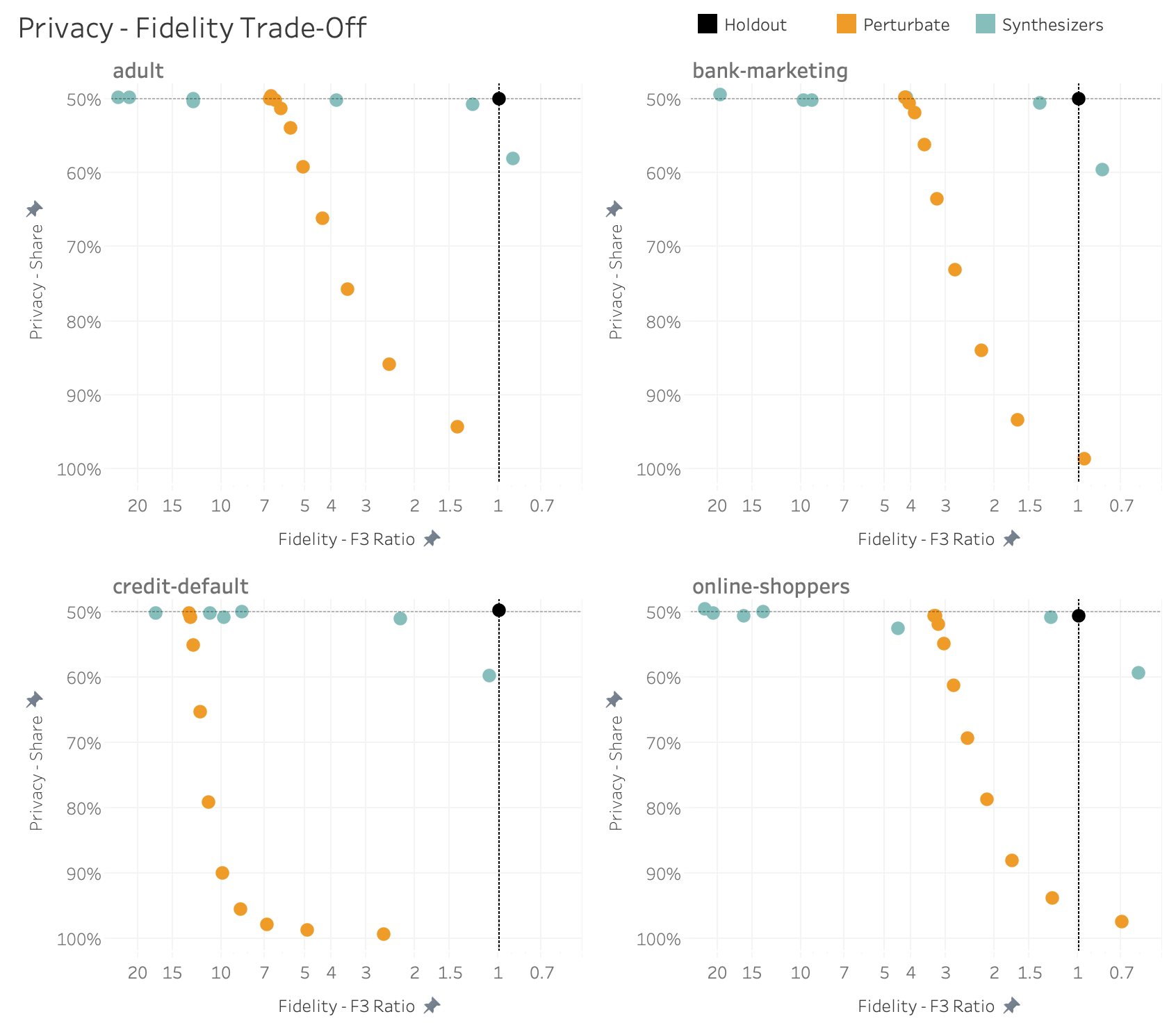}
\end{center}
\caption{Trade-off between privacy and fidelity for the presented empirical study}\label{fig:bench-tradeoff}
\end{figure}

Based on these results we can further visualize the uncovered empirical relationship between privacy and fidelity. The x-axes in Figure \ref{fig:bench-tradeoff} represent the three-way fidelity measure in relation to its corresponding value for the holdout dataset, i.e., $F^3(T,S)/F^3(T,H)$. The y-axes represent the reported share of records that are closer to training than to the holdout. Presented this way, the holdout dataset serves us as a ''north star'' for truly privacy-respecting data synthesizers in the upper right corner. The orange dots represent the range of perturbated datasets and reveal the difficulties of basic obfuscation techniques to protect privacy without sacrificing fidelity, particularly for higher-dimensional datasets. The turquoise marks on the other hand represent the performance metrics for a broad range of emerging synthesizers, whereas all except one exhibit DCR shares close to 50\%, and with some getting already very close to representing the characteristics of a true holdout dataset. 

\section{Discussion}\label{sec:discussion}

The field of supervised machine learning benefited from having commonly used benchmark datasets and metrics in place to measure performance across methods as well as progress over time. The emerging field of privacy-preserving structured synthetic data is still to converge onto commonly agreed fidelity and privacy measures, as well as to a set of canonical datasets to benchmark on. This research aims at complementing already existing methods by introducing a practical, assumption-free and easy-to-reason empirical assessment framework that can be applied for any black-box synthetization method and thus shall help to objectively capture and measure the progress in the field. In addition, the reported findings from the empirical benchmark experiments demonstrate the promise of AI-based data synthesis when compared to more traditional statistical disclosure techniques. However, they also highlight the need to not only assess fidelity but just as well the privacy risk of these newly emerging, powerful data generators. We hope that our research made a first step towards this direction.

\section*{Funding}

This research is supported by the "ICT of the Future” funding programme of the Austrian Federal Ministry for Climate Action, Environment, Energy, Mobility, Innovation and Technology.

\section*{Data Availability Statement}

A reference implementation of the framework, as well as all datasets of the empirical study are made available in a publicly accessible repository found at \href{https://github.com/mostly-ai/paper-fidelity-accuracy}{https://github.com/mostly-ai/paper-fidelity-accuracy}.

\bibliographystyle{arxiv}  
\bibliography{arxiv}

\end{document}